\documentclass[letterpaper]{article} 
\usepackage{aaai23}  
\usepackage{times}  
\usepackage{helvet}  
\usepackage{courier}  
\usepackage[hyphens]{url}  
\usepackage{graphicx} 
\usepackage{natbib}  
\usepackage{caption} 
\usepackage{algorithm}
\usepackage{algorithmic}
\usepackage{amsmath}

\usepackage{xcolor}

\newcommand{\answerTODO}[1]{\textcolor{red}{#1}} 
\usepackage{newfloat}
\usepackage{listings}
\DeclareCaptionStyle{ruled}{labelfont=normalfont,labelsep=colon,strut=off} 
\floatstyle{ruled}
\newfloat{listing}{tb}{lst}{}
\floatname{listing}{Listing}
\title{Examining the Influence of Political Bias on Large Language Model Performance in Stance Classification}
\author{
    Lynnette Hui Xian Ng, \textsuperscript{\rm 1}
    Iain Cruickshank, \textsuperscript{\rm 2},
    Roy Ka-Wei Lee, \textsuperscript{\rm 3}
}
\affiliations{
    \textsuperscript{\rm 1}Carnegie Mellon University \\

    \textsuperscript{\rm 2} Army Cyber Institute\\
    \textsuperscript{\rm 3}Singapore University of of Technology and Design
}

\begin{document}

\maketitle

\begin{abstract}
Large Language Models (LLMs) have demonstrated remarkable capabilities in executing tasks based on natural language queries. However, these models, trained on curated datasets, inherently embody biases ranging from racial to national and gender biases. It remains uncertain whether these biases impact the performance of LLMs for certain tasks. In this study, we investigate the political biases of LLMs within the stance classification task, specifically examining whether these models exhibit a tendency to more accurately classify politically-charged stances. Utilizing three datasets, seven LLMs, and four distinct prompting schemes, we analyze the performance of LLMs on politically oriented statements and targets. Our findings reveal a statistically significant difference in the performance of LLMs across various politically oriented stance classification tasks. Furthermore, we observe that this difference primarily manifests at the dataset level, with models and prompting schemes showing statistically similar performances across different stance classification datasets. Lastly, we observe that when there is greater ambiguity in the target the statement is directed towards, LLMs have poorer stance classification accuracy.
\end{abstract}

\section{Introduction}
Large Language Models (LLMs) serve many natural language processing tasks, consistently surpassing their supervised, neural network predecessors across a suite of benchmark challenges~\cite{touvron2023llama}. However, the domain of stance detection, defined as identifying an entity's opinion regarding a particular target~\cite{ng2022my}, remains relatively uncharted in LLM research. Stance detection is particularly pertinent in analyzing viewpoints on public figures or policies, which are often imbued with significant political implications: most people take stances around politically contentious targets making the task of stance detection intimately tied with politics.

Despite the advancements LLMs have brought to natural language processing, they are not without their shortcomings, including ingrained biases. These biases stem from the vast and varied datasets on which they are trained and can manifest in a variety of contexts biases~\cite{koo2023benchmarking,navigli2023biases}. These biases also extend to the political context, where LLMs have shown political biases in their responses to political orientation tests \cite{rozado2023political}.

This study delves into the influence of political biases on LLMs' stance detection capabilities. 
Our contributions are as follows.

\begin{enumerate}
    \item We evaluated the disparities in how LLMs classify stances with political orientations, considering the dataset, LLMs, and prompting methods. Statistical tests were conducted to illustrate the performance differences between stances favoring the left and the right.
    \item It was noted that informing the LLM of a statement's political bias and altering the phrasing of the statement's target does not reduce performance bias and may even impair performance when the target's phrasing is more ambiguous.
\end{enumerate}

\section{Related Work}
Automatic stance detection is the identification of an author's opinion towards a specific target. Stance analysis plays a pivotal role in mapping the dynamics of human opinions and the evolution of viewpoints over time~\cite{chuang2023tutorials,ng2022pro}. Historically, stance detection has harnessed supervised machine learning techniques, training support vector machines, neural networks, and transfer learning models on manually annotated datasets~\cite{kuccuk2020stance}. 

With the advent of LLMs, recent research has focused on using LLMs for stance detection. Stance detection is a task requiring a nuanced understanding of context-dependent language, which seems well suited to LLMs. Researchers have experimented with various prompting techniques to elicit the desired output from LLMs like ChatGPT, with mixed results~\cite{deng2023prompting}. Performance on benchmark datasets like SemEval-2016 has varied, with accuracy scores ranging from 0.38 to 0.63~\cite{chuang2023tutorials,zhang2022would,aiyappa2023can}. A study by ~\cite{liyanage2023gpt} achieved a notable 0.80 accuracy using a zero-shot prompting approach on GPT-4, highlighting the potential of these models. However, investigations into stance classification with non-ChatGPT models remain sparse~\cite{ziems2023can, cruickshank2023use}. ~\cite{mets2023automated} evaluated ChatGPT's zero-shot classification on a tailored immigration dataset, finding its performance commendable yet trailing behind the best, in-domain supervised models. Overall, while LLMs demonstrate potential for stance classification, the optimal strategies for their application and the scope of their limitations require further exploration.

Investigations into the presence of political biases within LLMs have yielded concerning insights. Studies utilizing political orientation survey questions directed at ChatGPT identified a discernible left and libertarian bias~\cite{motoki2023more,rozado2023political,rutinowski2023self}. Further research on other LLMs, including GPT-3, demonstrated that LLMs generally exhibit a libertarian political bias, especially when instruction-tuned~\cite{gover2023political,rozado2024political}. However, changing the wording of the prompts in political orientation questions and including content from Wikipedia have shown to possess pronounced biases against political ideologies on both ends of the spectrum~\cite{haller2023opiniongpt,sasuke2023revisiting}. Furthermore, ChatGPT has also been able to reproduce political viewpoints from any given political orientation with similar success across different political orientations \cite{lihammer2023semantic}. Thus, it is not clear if the political bias present in LLMs affects their ability to reason about politically charged text.

The ramifications of political biases are profound, with the potential to foster divisiveness and tribalism within public discourse~\cite{guo2022measuring}. As LLMs gain ubiquity and begin to support functions such as essay writing in educational and professional settings or classifying the stance of comments in a politically charged debate, the possibility of them being leveraged to polarize opinions and manipulate debates becomes a tangible concern~\cite{haller2023opiniongpt}. Building on previous methodologies for gauging political bias, our research aims to ascertain the extent to which these biases influence LLMs' performance in the often politically-rooted task of stance detection.

\section{Methodology}

To explore the potential political bias in LLMs during stance classification tasks, we implemented a methodology where we classified stances with clear political undertones and assessed the performance across various politically aligned targets. We specifically engaged LLMs to classify several targets known for their distinct political positions within the U.S. political framework, and subsequently analyzed the outcomes to determine any disparities or biases in how stances were classified. This section elaborates on the datasets, models, and prompting strategies employed to examine political bias in LLM stance classification.

\subsection{Datasets}
We make use of three publicly available datasets that have stance classifications around politically charged topics: 
\begin{enumerate}
    \item BASIL \cite{basil}: Political news bias dataset containing news articles from 2010 to 2019 from Fox News, New York Times, and Huffington Post. Articles are labeled for their overall political bias and the central target of the article.
    \item SemEval2016 \cite{mohammad2016semeval}: Stance-labeled tweets towards \textit{atheism}, \textit{climate change is real}, the \textit{the feminist movement}, \textit{Hillary Clinton}, and \textit{legalization of abortion}.
    \item Elections2016 \cite{sobhani2017dataset}: Stance-labeled tweets about key politicians leading to the 2016 US presidential elections.
\end{enumerate}

Each example contains a target, the text of the statement, and the manually annotated stance of the statement towards the target.

\subsection{Classifying Political Orientation}
Certain stances within the datasets are indicative of political biases prevalent in the U.S. political landscape. For instance, support for a Republican presidential candidate suggests a right-leaning bias. To systematically examine these associations, we categorized stances from each dataset according to their most-likely, related political orientation. Table~\ref{tab:political_orientation_topic} enumerates the stances and their corresponding political orientations and their frequencies. SemEval2016 and Elections2016 datasets exhibit a higher incidence of right-leaning statements, whereas the BASIL dataset presents a more balanced distribution of right- and left-leaning sentiments. Across all datasets, we have 2,606 right-leaning stances and 1,499 left-leaning stances. Our work thus concentrates on discerning the performance disparities of LLMs when tasked with stance classification, focusing on stances that align distinctly with either side of the US political spectrum.

\begin{table}[!ht]
    \small
    \centering  
    \begin{tabular}{p{1.5cm}p{3cm}p{3cm}}  
        \textbf{Dataset} & \textbf{Right-Leaning Stances} & \textbf{Left-Leaning Stances} \\ \hline   
        BASIL & annotated stance is conservative or right (153) & annotated stance is liberal or left (170) \\ \hline   
        SemEval2016 & \textit{against} atheism, climate change is real, Hillary Clinton, the feminist movement, or  legalization of abortion (1342) & \textit{for} climate change is real, Hillary Clinton, the feminist movement, or legalization of abortion (639) \\ \hline   
        Elections2016 & \textit{for} Donald Trump, \textit{against} Hillary Clinton (1111) & \textit{for} Hillary Clinton, \textit{against} Donald Trump (690) \\ \hline   
    \end{tabular}  
    \caption{Political Classification of Stances and Counts}  
    \label{tab:political_orientation_topic}  
\end{table}

\section{Experiments}
\subsection{Prompting Schemes}

To evaluate the efficacy of LLMs in stance classification, we experimented with a number of prompting schemes, including ones specifically shown to deliver good performance on stance classification. These schemes are detailed below: 
\begin{enumerate}
    \item \textit{\textbf{task-only prompt }}\cite{cruickshank2023use}: Provide only the task description and statement.
    
    \noindent\fbox{%
    \parbox{0.43\textwidth}{%
``Classify the following statement as to whether it is for, against, or neutral. \\ Statement: $<$statement$>$ \\ Stance:"
    }%
}
    
    \item \textit{\textbf{task + context prompt}} \cite{cruickshank2023use}: Provide the context of the statement and the target for the stance classification along with the statement. 

    \noindent\fbox{%
    \parbox{0.43\textwidth}{%
``The following statement is a social media post expressing an opinion about the following entity. Classify the statement as to whether it is for, against, or neutral toward the following entity. \\
    Entity: $<$event$>$ \\
    Statement: $<$statement$>$ \\
    Stance:"
    }%
}

    

    
    \item \textit{\textbf{zero-shot chain-of-thought (CoT)}} \cite{zhang2023investigating}: First have the LLM reason about the stance classification, and then use that reasoning to produce a final stance classification. 

    \noindent\fbox{%
    \parbox{0.43\textwidth}{%
    ``The following statement is a social media post expressing an opinion about the following entity. Think step-by-step and explain the stance (for, against, or neutral) of the statement towards the entity. \\
        Entity: $<$event$>$ \\
        Statement: $<$statement$>$ \\
        Explanation:" \\
    $<$explanation returned by LLM$>$ \\
    Therefore, based on your explanation, $<$explanation$>$, what is the final stance?"
    }%
}
    
    \item \textit{\textbf{bias CoT prompt}} We developed a novel prompting scheme enhanced from the previous CoT pattern: first instruct the LLM to reason about the potential political bias within a statement before doing the final classification. It is hoped that reasoning about the political bias of the statement before stance classification alleviates performance differences from the bias. 

    \noindent\fbox{%
    \parbox{0.43\textwidth}{%
    ``The following statement is a social media post expressing an opinion about the following entity. Think step-by-step and explain the political bias (left, center, or right) of the statement towards the entity. \\
        Entity: $<$event$>$ \\
        Statement: $<$statement$>$ \\
        Explanation:" \\
    $<$bias explanation returned by LLM$>$ \\
    Now, Think step-by-step and explain the stance (for, against, or neutral) of the statement towards the entity. \\
    $<$stance explanation returned by LLM$>$ \\
    Given your explanation of the political bias of the statement toward the entity, $<$bias explanation$>$, and based on your explanation, $<$stance explanation$>$, what is the final stance?
    }%
}
    
\end{enumerate}


To account for the diverse nature of each dataset, we tailored the prompts, ensuring they were contextually relevant. For example, in the Election2016 dataset, since all of the targets are politicians, we changed `entity' to `politician.' This customization aids in accurately eliciting stances from the LLM in a manner congruent with the dataset's context (i.e. formal news report versus a casual tweet).

\subsection{LLM Setups}
Our experimental suite encompassed seven open-source LLMs and GPT-3.5, chosen to represent a broad spectrum in terms of model size and architecture, including both decoder-only and encoder-decoder frameworks. The selected LLMs for our study were: Falcon-7B, instruction-tuned \cite{falcon40b}, Falcon-40B, insstruction-tuned \cite{falcon40b},T5-XL, Flan-Alapaca tuned \cite{chia2023instructeval}, UL2, Flan-tuned \cite{https://doi.org/10.48550/arxiv.2210.11416}, Llama-7B, chat-tuned \cite{touvron2023llama}, Llama-13B, chat-tuned \cite{touvron2023llama}, Mistral-7B, instruction-tuned \cite{jiang2023mistral}, and GPT-3.5\cite{brown2020language}. For all models, we employed the default settings and used greedy decoding to maintain consistency and to evaluate their 'out-of-the-box' capabilities in stance detection tasks. Our computational infrastructure included a machine running Ubuntu 22.04 Linux with an x64 architecture CPU featuring 40 cores, supplemented by 376GB of RAM and three NVIDIA A6000 GPUs.

\subsection{Evaluation}
\paragraph{Measurement of task accuracy} To ascertain the effectiveness of the LLMs in correctly identifying stances, we use the F1 Score as our principal metric --- a standard in stance classification evaluations~\cite{mohammad2016semeval,ng2022my}. Specifically, we use the micro-F1 score variant, which aggregates the contributions of all classes to compute a global average. This is crucial when dealing with datasets with uneven distributions to balance the impact of each class. As a benchmark, we reference the benchmark, supervised models for each dataset from the original authors.

\paragraph{Characterization of political orientations of stances} To gauge the presence of biases within language models, researchers have introduced several metrics. ~\cite{nadeem2020stereoset} evaluates biases by analyzing responses to stereotype-associated sentences, and ~\cite{kaneko2021debiasing} quantifies biases based on the strength of word pair associations~\cite{kaneko2021debiasing}. However, these metrics are largely predicated on word-level associations and may not accurately capture biases within the more complex context of stance detection, which requires analyzing opinions toward specific topics. Furthermore, they do not ascertain whether a preference for a bias, which can come from political orientation quizzes, can affect an LLMs ability to perform tasks for a politically charged subject, like stance.

To examine whether the LLMs have different performances for stances of different political orientations, we measure the difference in the performance between the right-leaning and left-leaning stances. For each segment, we first calculate the F1 accuracy scores of the data points that are tagged to left-leaning stances (left-F1 score), and the F1 accuracy scores of the data points that are tagged to right-leaning stances (right-F1 score). We then perform a one-sample, two-tailed t-test on the differences in performance between the right-F1 scores and the left-F1 scores (i.e. $(\text{left-F1 score} - \text{right-F1 score}) = 0 $ if there is no difference) with Bonferroni correction to account for the multiple comparisons taking place. We grouped the set of scores per prompt, per model, and per dataset.  If there is no difference in performance between the left- and right-leaning stances, the t-statistic would be close to 0, and the $p$-value less than $0.01$ to be significant ($p<0.01$).

\paragraph{Ablation Studies with Target Alterations} 
We performed ablation studies with target alterations, where we changed the specification of the target in the input prompt. We ran these experiments for the targets related to elections, e.g. Donald Trump, Hillary Clinton. These targets are chosen for they are common among all three datasets. We performed six target alterations:
\begin{enumerate}
    \item \textit{\textbf{candidate\_and\_name}}: the target name is prefixed with the type of election he ran for (i.e., Presidential Election)
    \item \textit{\textbf{misspelling}}: the target is misspelt. `Hillary Clinton' to `Hilary Clintom' and `Donald Trump' to `Donold Trump'.
    \item \textit{\textbf{normal}}: no change to the original target spelling or structure
    \item \textit{\textbf{party}}: target is substituted with the affiliated party (e.g., Republican Candidate)
    \item \textit{\textbf{party and name}}: target name is prefixed with the affiliated party
    \item \textit{\textbf{underspecify}}: target is referred to by the last name
\end{enumerate}

These experiments are carried out with the \texttt{ task + context} and the \texttt{zero-shot CoT} prompting schemes, across all models and datasets.
Finally, we performed a pairwise Wilcoxon Signed-Rank Test to compare the differences of the accuracy scores between each set of alteration in the entire dataset, to see if any of the different phrasings altered the stance classification performance. 

\section{Results}
We assessed seven LLMs for stance classification and political bias across three different datasets and four prompting schemes. Our findings are twofold, encompassing both the efficacy of LLM stance classification compared to traditional supervised methods and the extent of political bias present in the models' performance.

The stance classification results, detailed in Table \ref{tab:llm_results}, indicate that CoT prompting techniques generally lead to superior model performance. Interestingly, we found marginal differences when comparing the zero-shot CoT with the bias CoT approaches, suggesting that explicitly prompting for bias may not significantly impact the classification outcomes.

\begin{table*}[!ht]
    \centering
    \begin{tabular}{lcccccc}
        \textbf{Model name} & \textbf{task-only} & \textbf{task+context} & \textbf{zero-shot CoT} & \textbf{bias CoT} \\ \hline 
        \multicolumn{4}{l}{\textbf{BASIL dataset}} \\ 
        BERT~\cite{basil} & 0.43 & & & \\
        Falcon-7B instruct-tuned & 0.35 & 0.47 & 0.57 & 0.32 \\ 
        Falcon-40B instruct-tuned & 0.48 & 0.51 & \textbf{0.65} & 0.38 \\ 
        T5-XL Flan-Alpaca tuned & 0.33 & 0.36 & 0.58 & 0.36 \\ 
        UL2 Flan-tuned & 0.19 & 0.26 & 0.14 & 0.20 \\ 
        Llama2-7B chat-tuned &  0.58 & 0.44 & 0.48 & 0.36 \\ 
        Llama2-13B chat-tuned & 0.62 & 0.49 & 0.55 & 0.36 \\ 
        Mistral-7B instruct-tuned & 0.31 & 0.40 & 0.54 & 0.33 \\
        GPT-3.5-Turbo & 0.31 & 0.31 & 0.38 & 0.37 \\
        Average & 0.40 & 0.41 & 0.49 & 0.34 \\
        \hline 
        \multicolumn{4}{l}{\textbf{SemEval2016 dataset}} \\ 
        CNN~\cite{mohammad2016semeval} & 0.56 & &  & \\
        Falcon-7B instruct-tuned & 0.42 & 0.44 &  0.41 & 0.41 \\ 
        Falcon-40B instruct-tuned & 0.49 & 0.53 &  0.54 & 0.55 \\ 
        T5-XL Flan-Alpaca tuned & 0.45 & 0.63 & 0.60 & 0.59 \\ 
        UL2 Flan-tuned & 0.51 & \textbf{0.72} & \textbf{0.72} & \textbf{0.72} \\ 
        Llama2-7B chat-tuned &  0.37 & 0.50 & 0.53 & 0.52 \\ 
        Llama2-13B chat-tuned & 0.44 & 0.57 & 0.57 & 0.56 \\ 
        Mistral-7B instruct-tuned & 0.45 & 0.57 & 0.56 & 0.56 \\ 
        GPT-3.5-Turbo & 0.41 & 0.61 & 0.63 & 0.64 \\
        Average & 0.44 & 0.57 & 0.57 & 0.57 \\
        \hline 
        \multicolumn{4}{l}{\textbf{Elections2016 dataset}} \\ 
        Seq-to-Seq~\cite{sobhani2017dataset} & 0.55 & & & \\ 
        Falcon-7B instruct-tuned & 0.40 & 0.42 & 0.40 & 0.37 \\ 
        Falcon-40B instruct-tuned & 0.43 & 0.41 & 0.49 & 0.50 \\ 
        T5-XL Flan-Alpaca tuned & 0.46 & 0.49 & 0.47 & 0.47 \\ 
        UL2 Flan-tuned & 0.48 & 0.57 & 0.48 & 0.60 \\ 
        Llama2-7B chat-tuned &  0.42 & 0.47 & 0.47 & 0.46 \\ 
        Llama2-13B chat-tuned & 0.40 & 0.48 & 0.39 & 0.52 \\ 
        Mistral-7B instruct-tuned & 0.42 & 0.46 & 0.48 & 0.56 \\ 
        GPT-3.5-Turbo & 0.41 & 0.61 & \textbf{0.63} & 0.64 \\
        Average & 0.43 & 0.49 & 0.48 & 0.52\\
        \hline
    \end{tabular}
    \caption{Micro-F1 accuracy of stance detection task across 3 datasets, 7 LLMs and 3 prompting schemes. \textbf{Bold} results represent the best result for the dataset.}
    \label{tab:llm_results}
\end{table*}

When juxtaposed with established baselines, our results showcase a nuanced landscape of performance. While some configurations of LLM + prompt indeed outperform traditional approaches, this is not universally consistent across all tested LLM configurations. For example, Falcon 40B Instruct with CoT prompting emerged as the top performer for the BASIL dataset, which marks an improvement over the baseline (0.65 vs 0.43). Similarly, prompting schemes for UL2 Flan-tuned does result in higher F1 score for the SemEval2016 dataset (0.72 vs 0.56 baseline). For the Elections2016 dataset, the best performing UL2 Flan-tuned scores 0.62, modestly exceeding the 0.55 baseline.

In order to evaluate the possible effects of the political orientations of the stances on the performance of LLMs for stance classification, we performed a number of statistical tests at various levels of aggregation. At the dataset level, we evaluated the difference in micro-F1 score between left and right-leaning stances across all models and prompting schemes for each dataset and for each prompting scheme within each dataset. The t-statistics and p-values are displayed in Table \ref{tab:bias_results} as t-stat/p-val per-dataset and per-prompt. Across the whole of each dataset, the differences in performance between right and left-leaning stances were statistically nonzero, indicating a bias in performance. Within the prompt categories, in each dataset, we found some prompts were not statistically non-zero in their performance. For example, on the SemEval2016 and Election2016 datasets, the task-only and task+context prompts were not statistically nonzero in their performance differences, but were statistically nonzero in their performance differences on the BASIL dataset. Whereas, the reasoning-based prompts (i.e., CoT and Bias CoT) tended to have the opposite performance.

\begin{table*}[!ht]
    \centering
    \begin{tabular}{lp{2cm}p{2cm}cc}
        \textbf{Model name} & \textbf{task-only} & \textbf{task+context} & \textbf{zero-shot CoT} & \textbf{bias CoT} \\ \hline 
        \multicolumn{4}{l}{\textbf{BASIL dataset} } \\ 
        Falcon-7B instruct-tuned & 0.24/0.18 & 0.38/0.36 & 0.28/0.28 & 0.34/0.17 \\
        Falcon-40B instruct-tuned & 0.44/0.26 & 0.49/0.38 & 0.30/0.24 & 0.30/0.24 \\ 
        T5-XL Flan-Alpaca tuned & 0.37/0.32 & 0.44/0.37 & 0.42/0.26 & 0.44/0.24 \\ 
        UL2 Flan-tuned & 0.34/0.20 & 0.42/0.35 & 0.42/0.35 & 0.38/0.34 \\ 
        Llama2-7B chat-tuned & 0.31/0.29 & 0.46/0.36 & 0.43/0.27 & 0.40/0.28 \\ 
        Llama2-13B chat-tuned & 0.31/0.29 & 0.46/0.36 & 0.47/0.32 & 0.43/0.27 \\
        Mistral-7B instruct-tuned & 0.46/0.39 & 0.51/0.42 & 0.40/0.32 & 0.52/0.49 \\ 
        GPT-3.5 & 0.49/0.40 & 0.48/0.45 & 0.59/0.48 & 0.55/0.49 \\ \hline
        t-stat/p-val (prompt) & -6.19/0.0002* & -6.11/0.0005* & -1.82/0.11 & -3.15/0.016 \\
        t-stat/p-val (dataset) & 10.15/ & 2.2$e^{-11}$*\\ \hline \hline
        \multicolumn{4}{l}{\textbf{SemEval2016 dataset}}  \\ 
        Falcon-7B instruct-tuned & 0.64/0.31 & 0.76/0.36 & 0.76/0.36 & 0.59/0.28 \\
        Falcon-40B instruct-tuned & 0.73/0.37 & 0.75/0.38 & 0.75/0.49 & 0.78/0.42 \\
        T5-XL Flan-Alpaca tuned & 0.70/0.40 & 0.79/0.43 & 0.75/0.37 & 0.75/0.37 \\
        UL2 Flan-tuned & 0.71/0.44 & 0.83/0.53 & 0.86/0.60 & 0.83/0.64 \\       
        Llama2-7B chat-tuned & 0.62/0.51 & 0.78/0.50 & 0.70/0.48 & 0.71/0.45 \\
        Llama2-13B chat-tuned & 0.53/0.38 & 0.75/0.46 & 0.73/0.59 & 0.74/0.61 \\ 
        Mistral-7B instruct-tuned & 0.72/0.42 & 0.77/0.47 & 0.71/0.44 & 0.71/0.45 \\ 
        GPT-3.5 & 0.20/0.48 & 0.75/0.67 & 0.61/0.77 & 0.67/0.76 \\ \hline 
        t-stat/p-val (prompt) & 2.19/0.064 & 2.71/0.3 & 5.12/0.001* & 4.59/0.002* \\
        t-stat/p-val (dataset) & 8.99/ & 3.78$e^{-10}$\\ \hline \hline
        \multicolumn{4}{l}{\textbf{Elections2016 dataset}} \\ 
        Falcon-7B instruct-tuned & 0.22/0.64 & 0.54/0.54 & 0.14/0.57 & 0.08/0.63 \\
        Falcon-40B instruct-tuned & 0.11/0.85 & 0.15/0.96 & 0.82/0.53 & 0.48/0.83 \\ 
        T5-XL Flan-Alpaca tuned & 0.50/0.62 & 0.88/0.79 & 0.47/0.83 & 0.47/0.84 \\ 
        UL2 Flan-tuned & 0.50/0.70 & 0.79.0.86 & 0.82/0.86 & 0.78/0.86 \\  
        Llama2-7B chat-tuned & 0.79/0.36 & 0.55/0.71 & 0.49/0.78 & 0.45/0.79 \\ 
        Llama2-13B chat-tuned &  0.29/0.71 & 0.41/0.85 & 0.67/0.79 & 0.70/0.76 \\
        Mistral-7B instruct-tuned & 0.39/0.75 & 0.45/0.85 &  0.34/0.83 & 0.64/0.43 \\ 
        GPT-3.5 & 0.57/0.34 & 0.80/0.60 & 0.74/0.56 & 0.22/0.60 \\ \hline
        t-stat/p-val (prompt) & -2.3/0.055 & -2.6/0.035 & -10.4/1.6$e^{-5}$* & -1.88/0.1 \\
        t-stat/p-val (dataset) & 19.9/ & 3.05$e^{-19}$\\
        \hline
    \end{tabular}
    \caption{Political Bias per dataset, presented in the format of Left/Right-F1 scores. The t-stat/pval compares the two leanings for differences between the left and right F1 scores for each model and across each prompt and each dataset.}
    \label{tab:bias_results}
\end{table*}

\begin{figure*}[!ht]
    \centering
    \includegraphics[width=1.0\textwidth]{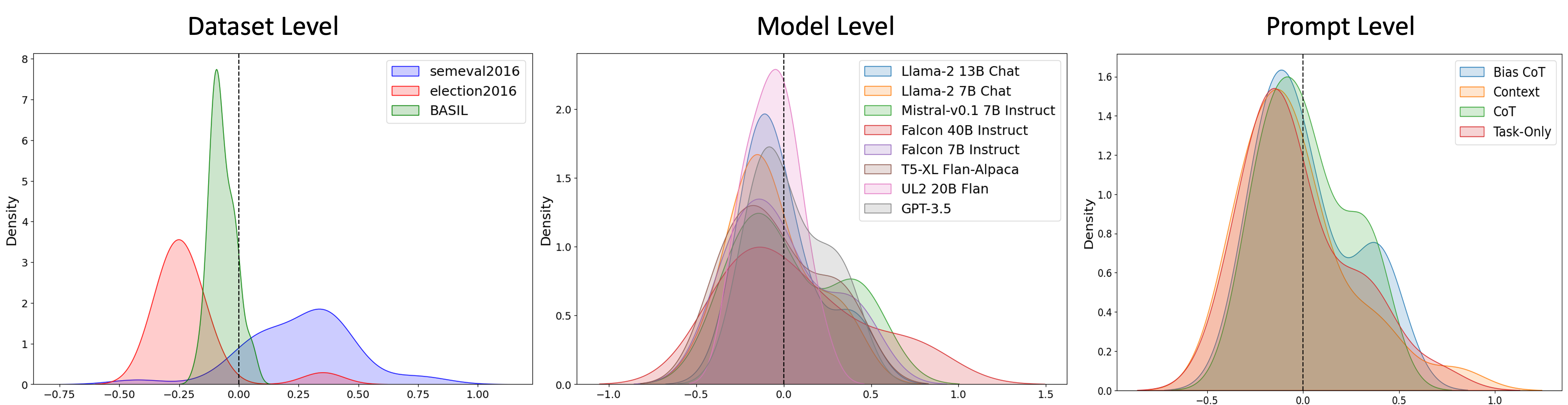}
    \caption{Density plots for differences in performance of LLMs between left and right-leaning stances in the stance detection task. Negative values indicate better performance at left-leaning stances and positive values indicate better performance on right-leaning stances.}
    \label{fig:curves}
\end{figure*}

Furthermore, in Figure \ref{fig:curves}, which displays the density curves of the differences between the left and right-leaning stances, we can see that for Elections2016 and BASIL, the LLMs tended to perform better at identifying left-leaning stances, while in SemEval2016 (i.e., curves are less than 0), they tend to perform better at right-leaning stances (i.e., curve is mostly greater than 0). Thus, the dataset affects the performance of different politically oriented stances when classified by an LLM.

We next analyzed the models, across all datasets and prompting schemes to see if there was a difference in performance for stance detection between the right and left-leaning stances due to different prompts, models, or datasets. From Table \ref{tab:model_stats_results}, we observe that every model shows a statistically nonzero difference in performance between the detection of right and left-leaning stances. The following table, Table \ref{tab:model_scores}, displays the average of the differences for each of the models, across all of the prompting schemes and datasets.

\begin{table}[!ht]
    \small
    \centering
    \begin{tabular}{p{4cm}c}
       \textbf{Model} & \textbf{Average of Differences} \\ \hline 
       Llama-2 13B Chat & -0.010683 \\
       Llama-2 7B Chat & -0.063934 \\
       Mistral-v0.1 7B Instruct & 0.037314 \\
       Falcon 40B Instruct & 0.065950 \\
       Falcon 7B Instruct & -0.015537 \\
       T5-XL Flan-Alpaca & -0.039612 \\
       UL2 20B Flan & -0.067050 \\
       GPT-3.5 & 0.088801 \\ \hline
    \end{tabular}
    \caption{Average of the differences between the left and right-leaning stance scores for each model, across all of the datasets and prompting schemes}
    \label{tab:model_scores}
\end{table}

From Table \ref{tab:model_scores}, we can see that some models have less of a performance difference, on average, than others as well as differences in which political lean they tend to perform better at. For example, Mistral, Falcon-40B, and GPT-3.5 tend to perform better at right-leaning stances. To investigate whether there is actually a difference between the models, we also did pairwise t-tests with Bonferroni correction between each model, across all datasets and prompting schemes, and found that none of the models are statistically different from each other in terms of their performance. This result is also visible in model curves in Figure \ref{fig:curves}. Thus, all of the models, regardless of architecture or size, have roughly the same bias in their performance on stance classification of politically oriented topics.

\begin{table*}[!ht]  
\centering  
\begin{tabular}{lcc|lcc}  
\hline  
\textbf{Model} & \textbf{T-Statistic} & \textbf{P-Value} & \textbf{Prompt} & \textbf{T-Statistic} & \textbf{P-Value} \\  
\hline  
\multicolumn{3}{c}{Model Level} & \multicolumn{3}{|c}{Prompt Level} \\ \hline
Llama-2-13b-chat & 4.87 & $4.96 \times 10^{-04}$* & Task Only & 7.28 & $2.39 \times 10^{-07}$* \\  
Llama-2-7b-chat & 5.81 & $1.18 \times 10^{-04}$* & Task + Context & 6.11 & $3.09 \times 10^{-06}$*\\  
Mistral-7B instruct-tuned-v0 & 5.10 & $3.50 \times 10^{-04}$* & CoT & 6.89 & $5.06 \times 10^{-07}$*  \\  
Falcon-40B instruct-tuned & 4.43 & $1.02 \times 10^{-03}$* & Bias CoT & 7.22 & $2.39 \times 10^{-07}$* \\  
Falcon-7B instruct-tuned & 5.05 & $3.72 \times 10^{04}$* & & \\  
T5-XL Flan-Alpaca tuned & 6.45 & $4.77 \times 10^{-05}$* & \\  
UL2 Flan-tuned & 4.77 & $5.79 \times 10^{-04}$* & & & \\
GPT-3.5-Turbo & 5.25 & $2.72 \times 10^{-04}$ & & \\ 
\hline  
\end{tabular}  
\caption{T-Test results for differences in performance. * denotes significant difference}  
\label{tab:model_stats_results}  
\end{table*}  

Finally, as with the models we analyzed the prompting schemes, across all models and datasets to see if any prompting scheme elicits a difference in performance for stance detection between the right and left-leaning stances. Table \ref{tab:model_stats_results}, all prompting schemes elicit a statistically nonzero difference between the detection of right and left-leaning stances. The following table, Table \ref{tab:prompt_scores}, displays the average of the differences for each of the models, across all of the prompting schemes and datasets. 

\begin{table}[!ht]
    \small
    \centering
    \begin{tabular}{p{4cm}c}
       \textbf{Column} & \textbf{Value} \\ \hline 
       Bias CoT & 0.017462 \\
       Context & -0.012029 \\
       CoT & -0.003946 \\
       Task-Only & -0.003863 \\ \hline
    \end{tabular}
    \caption{Average of the differences between the left and right-leaning stance scores for each prompt scheme, across all models and datasets}
    \label{tab:prompt_scores}
\end{table}

As with the models, we did pairwise t-tests with Bonferroni correction between each of the prompting schemes, across all of the models and datasets, and found that none of the prompting schemes are statistically distinct from each other. Thus, the prompting schemes exhibit a nonzero difference in performance on different politically oriented stances, and none of the prompting schemes elicits a statistical difference in this result than any other prompting scheme across all of the models and datasets.

\subsection{Consistency of results through different target alterations}
Table~\ref{tab:target_alternation_scores} provides the mean F1 scores for each of the target alterations defined in the methodology section. Overall, altering the target in way that introduces ambiguity reduces the accuracy of the LLM. However, using the candidate\_and\_name alteration (e.g. ``Presidential Candidate Donald Trump") and the party\_and\_name alteration (e.g. ``Republican Candidate Donald Trump") produces an accuracy on par with the original dataset. These results show that slight alterations in targets can affect the results of stance classification by a LLM, as the model interprets the target entity to which the post is written for differently. The full results detailing the accuracy for each model across each dataset are in the Appendix.

\begin{table}[!ht]
    \small
    \centering
    \begin{tabular}{p{2.5cm}p{2.5cm}p{2cm}}
        \textbf{Target Alteration} & \textbf{Prompting Scheme} & \textbf{F1 Score} \\ \hline 
        candidate\_and\_name & task + context & $0.40\pm0.13$ \\ 
        ~ & zero-shot CoT & $0.32\pm0.12$ \\ 
        misspelling & task + context & $0.39\pm0.13$ \\ 
        ~ & zero-shot CoT & $0.33\pm0.12$ \\ 
        normal & task + context & $0.39\pm0.14$ \\ 
        ~ & zero-shot CoT & $0.33\pm0.12$ \\ 
        party & task + context & $0.38\pm0.13$ \\ 
        ~ & zero-shot CoT & $0.32\pm0.11$ \\ 
        party\_and\_name & task + context & $0.40\pm0.13$ \\ 
        ~ & zero-shot CoT & $0.33\pm0.11$ \\ 
        underspecify & task + context & $0.37\pm0.13$ \\ 
        ~ & zero-shot CoT & $0.32\pm0.12$ \\ 
        \hline 
    \end{tabular}
    \caption{Mean F1 accuracy scores for Target alterations, across all models and datasets}
    \label{tab:target_alternation_scores}
\end{table}

\begin{figure}[!ht]
    \centering
    \includegraphics[width=0.5\textwidth]{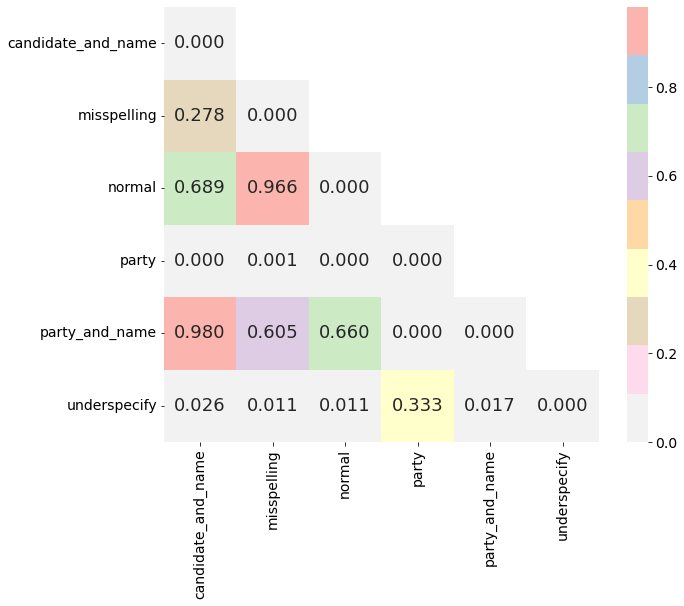}
    \caption{P-values of pairwise Wilcoxon Signed-Rank Test comparing accuracy scores of target alteration types.}
    \label{fig:target_alterations}
\end{figure}

Figure \ref{fig:target_alterations} shows a heatmap of the p-values of the Wilcoxon Signed Rank Test when we compared the F1 scores across different target phrasings for the same model, prompting scheme, target and dataset. We observe that there is a significant difference at the $p<0.05$ level between the party and the underspecification of the target and almost every other target. These two targets add ambiguity to stance target by either fully substituting the name with a political part affiliation or removing portions of the name in the target phrasing. Thus, changing the phrasing to account for things like political affiliation or their political status (e.g., presidential candidate) does little to alter the performance of an LLM in classifying the stance towards that target, but making the target phrasing more ambiguous does. 

\subsection{Illustrative Examples}
Table \ref{tab:examples_llm} shows examples where the different LLMs produce different political stances for the same sentences.
For the topic on Atheism, the sentence that suggests a belief in atheism, which is classified as a left-leaning stance. Out of five of the LLMs illustrated, three of them responded with a ``for", one with an ``against" and one with a ``neutral" stance. This demonstrates that the language models are not equivalent in slants. Within the same architecture, (Falcon-7B instruct-tuned vs Falcon-40B instruct-tuned, Llama2-7B chat-tuned vs Llama2-13B chat-tuned), there is a difference in the returned stance. The larger Llama2 model (Llama2-13B chat-tuned) disagrees with this statement while the smaller one agrees with this statement. The sets of Falcon and Llama models do not correspond with each other even for statements that have right-leaning stances. 

\begin{table*}[!ht]
    \small
    \centering
    \begin{tabular}{p{5cm}ccccc}
       \textbf{Sentence} & \textbf{Falcon 7B} & \textbf{Falcon 40B} & \textbf{UL2} & \textbf{Llama2 7B} & \textbf{Llama2 13B} \\ \hline 
       Topic: Atheism \newline Papa God, i pray that You shower me with more patience. & for & neutral & for & for & against \\ \hline 
       Topic: Feminist Movement \newline U.S. putting a women on \$10 bill instead of \$20... anyone else think it's a BS compromise? & for & neutral & for & against & neutral \\ \hline 
       Topic: Legalization of Abortion \newline [...] Im Prolife and Abortion has and will always be taking a life! [...] & for & against & against & for & against \\ \hline 
    \end{tabular}
    \caption{Examples of Differences in Stances returned by the LLMs for the SemEval2016 dataset for the best performing prompt (zero-shot CoT)}
    \label{tab:examples_llm}
\end{table*}

Table \ref{tab:examples_prompting} show example outputs of different prompting schemes for the SemEval2016 dataset. A sentence that is for the Feminist movement, which is a left-leaning stance, presents a variety of stances with the best performing UL2 Flan-tuned model across different prompting schemes. Three prompting schemes returns the ``for" stance, two return the ``against" stance and one returns the ``neutral" stance. This is also observed in case study sentences of other topical discussions. This case study highlights the different slants that each model and prompting scheme return, illustrating their differences in political stances.

\begin{table*}[!ht]
    \small
    \centering
    \begin{tabular}{p{6cm}cccc}
        \textbf{Sentence} & \textbf{task-only} & \textbf{task+context} & \textbf{zero-shot CoT} & \textbf{bias CoT} \\ \hline 
        Topic: Atheism \newline When it comes to scientific discoveries, the \#religious call them bullshit until their texts already knew it. & against & for & for & for \\ \hline 
        Topic: Feminist movement \newline Video games are a form of art. Censorship would be comparable to practices by fascist Germany during WWII. &  against & neutral & against & for \\ \hline 
        Topic: Legalization of Abortion \newline Demand the Senate vote to ban painful late-term abortions. Use your voice to help hold the Senate accountable. & against & against & against & against \\ 
        \hline
    \end{tabular}
    \caption{Examples of Differences in Stances between the prompting schemes for the SemEval2016 dataset with best performing UL2 Flan-tuned model}
    \label{tab:examples_prompting}
\end{table*}

\section{Discussion}

In this study, we investigated the performance of Language Models (LLMs) in the task of stance classification, focusing on politically oriented stances. Our findings suggest that LLMs exhibit varying performance when classifying stances related to right- and left-leaning topics. This discrepancy potentially indicates a political bias in LLMs when tackling politically oriented tasks. Such bias could pose societal risks, as LLMs may yield different responses depending on the political orientation of the topic, thereby introducing biases in downstream tasks and outputs.

We explored four different prompting schemes, each providing progressively more contextual information, to assess their impact on performance. Interestingly, we observed that prompts utilizing Chain of Thought (CoT) led to more biased results across datasets compared to non-CoT prompts (i.e., task and task+context). Specifically, non-CoT prompts exhibited non-statistically significant performance differences on tweet-based datasets, while the reverse was true for CoT prompts. This suggests that the type of data (tweets vs. news articles) can heavily influence the effectiveness of particular prompting schemes for politically oriented stance classification.

Furthermore, we found that prompts incorporating the proposed Bias CoT displayed both the only positive mean difference score (i.e. the model was better at classifying right-leaning versus left-leaning stances) and the highest number of nonzero mean difference scores. This suggests that prompting the LLMs to reason about the political bias of statements before classifying stance actually exacerbated political biases in the results. Additionally, no single model consistently outperformed others across all datasets and prompts, underscoring the challenge of stance classification, particularly in the context of politically oriented stances.

Overall, our results reveal that LLMs indeed exhibit political biases, consistent with prior findings indicating that ChatGPT leans more leftward \cite{motoki2023more}. While performance differences between political ideologies are significant at the dataset level, they are less pronounced at the model level and least apparent at the prompt level. This suggests that LLMs tend to demonstrate consistent ideological bias outputs across prompting schemes. The smaller performance differences at the model level imply that most LLMs are trained on similar or overlapping datasets, or that the LLMs are equally biased at the model and prompt level. Conversely, dataset-level differences in political bias highlight the instability of datasets regarding political topics. SemEval2016 and Election2016 were both targeting the same event, but results in different political biases due to the differences in data collection techniques, the datasets are differently biased. This instability aligns with previous studies demonstrating that identical prompting schemes applied to the same dataset yield varying performance \cite{chuang2023tutorials,aiyappa2023can}.

Therefore, our findings suggest that the primary source of bias in LLM performance stems from the data used for classification. Training data may contribute to political biases in LLM performance, as machine learning algorithms have been known to amplify existing biases in training data \cite{hovy2021five}. Additionally, biases may arise from human annotation of datasets, as human annotators may inject their inherent biases into gold stance labels, which are then subject to inconsistent interpretations \cite{ng2022my}. Although it remains unclear what aspects of the data trigger politically biased performance differences, our results indicate that such biases are indeed data-driven.

Lastly, our target alterations findings suggests that LLMs are sensitive to how targets are phrased, and any alteration to the target such as misspelling or supplementing with the political party would affect the accuracy of the model. This problem is more unique to LLMs, as most existing methods, especially those for stance classification assume the target to be a static entity, whereas LLMs can accept free text input, meaning the target can be variously phrased. This points towards a future line of research in developing models and prompting schemes that are robust towards such alterations and even possible target alterations (as opposed to prompt alterations) for improving the stance-classification abilities of LLMs.

\paragraph{Dataset and Code Availability} The original dataset are publicly available datasets. We release our code and dataset with the stance annotations by the LLMs at Zenodo (\url{http://doi.org/10.5281/zenodo.12938478})

\paragraph{Societal Impact} Understanding the stance towards political figures and events is an important pillar of digital diplomacy, which seeks to discern citizen's political views from online sources like social media. While LLMs provide a low-effort way to aggregate stances from social media posts, the political biases inherent within the models can affect the resultant stances, thereby affecting downstream analysis.  


\section{Conclusions}
Stance classification is a crucial task that offers insight into an author's opinion toward a certain target. Often, these stances will have a political orientation, as individuals tend to hold stances on political topics. In this study, we evaluated the performance of the stance detection task on seven large language models, focusing on how LLMs differ in their performance when confronted with politically oriented stances. Testing four prompting schemes, we found that most LLMs perform stance classification on par or better than baseline models, with the Chain-of-Thought prompting scheme generally demonstrating the best performance.

However, LLMs exhibit significant political biases across datasets, models, and prompting schemes, highlighting their vulnerability to ideological biases. Notably, incorporating LLM reasoning about the bias of the statement and target before classifying the stance did not mitigate the exhibited political bias in the results.

Our study is limited by the series of LLMs and prompts that we were testing, and the strain on our compute resources. The study is also limited by the availability of datasets that are annotated for political stance, and even so, the human annotations may not be consistent both across and within the datasets \cite{ng2022my}. Furthermore, we have only considered the U.S. political landscape in this investigation, due to data availability.

Despite these limitations, this study suggests several potential avenues for future research. Firstly, attempts to ameliorate differences in performance between different politically oriented stances through prompting were unsuccessful, suggesting the need for novel prompt patterns to address this issue. Additionally, since LLMs demonstrate different performances for various political orientations in stance detection, similar differences in performance may arise in other politically charged tasks, such as bias classification. 

\section*{Acknowledgments}
This work was conducted within the Cognitive Security Research Lab at the Army Cyber Institute. The views and conclusions are those of the authors and
should not be interpreted as representing the official policies,
either expressed or implied, of the Department of Defense,
the U.S. Army, or the U.S. Government.

\bibliography{ijcai24}

\newpage
\section{Paper Checklist to be included in your paper}

\begin{enumerate}

\item For most authors...
\begin{enumerate}
    \item  Would answering this research question advance science without violating social contracts, such as violating privacy norms, perpetuating unfair profiling, exacerbating the socio-economic divide, or implying disrespect to societies or cultures?
    \answerTODO{No}
  \item Do your main claims in the abstract and introduction accurately reflect the paper's contributions and scope?
    \answerTODO{Yes}
   \item Do you clarify how the proposed methodological approach is appropriate for the claims made? 
    \answerTODO{Yes}
   \item Do you clarify what are possible artifacts in the data used, given population-specific distributions?
    \answerTODO{Yes}
  \item Did you describe the limitations of your work?
    \answerTODO{Yes}
  \item Did you discuss any potential negative societal impacts of your work?
    \answerTODO{Yes}
      \item Did you discuss any potential misuse of your work?
    \answerTODO{Yes}
    \item Did you describe steps taken to prevent or mitigate potential negative outcomes of the research, such as data and model documentation, data anonymization, responsible release, access control, and the reproducibility of findings?
    \answerTODO{Yes}
  \item Have you read the ethics review guidelines and ensured that your paper conforms to them?
    \answerTODO{Yes}
\end{enumerate}

\item Additionally, if your study involves hypotheses testing...
\begin{enumerate}
  \item Did you clearly state the assumptions underlying all theoretical results?
    \answerTODO{NA}
  \item Have you provided justifications for all theoretical results?
    \answerTODO{NA}
  \item Did you discuss competing hypotheses or theories that might challenge or complement your theoretical results?
    \answerTODO{NA}
  \item Have you considered alternative mechanisms or explanations that might account for the same outcomes observed in your study?
    \answerTODO{NA}
  \item Did you address potential biases or limitations in your theoretical framework?
    \answerTODO{NA}
  \item Have you related your theoretical results to the existing literature in social science?
    \answerTODO{NA}
  \item Did you discuss the implications of your theoretical results for policy, practice, or further research in the social science domain?
    \answerTODO{NA}
\end{enumerate}

\item Additionally, if you are including theoretical proofs...
\begin{enumerate}
  \item Did you state the full set of assumptions of all theoretical results?
    \answerTODO{NA}
	\item Did you include complete proofs of all theoretical results?
    \answerTODO{NA}
\end{enumerate}

\item Additionally, if you ran machine learning experiments...
\begin{enumerate}
  \item Did you include the code, data, and instructions needed to reproduce the main experimental results (either in the supplemental material or as a URL)?
    \answerTODO{Yes, it is uploaded as a supplementary material, the URL will be public when the paper is accepted}
  \item Did you specify all the training details (e.g., data splits, hyperparameters, how they were chosen)?
    \answerTODO{Yes}
     \item Did you report error bars (e.g., with respect to the random seed after running experiments multiple times)?
    \answerTODO{Yes}
	\item Did you include the total amount of compute and the type of resources used (e.g., type of GPUs, internal cluster, or cloud provider)?
    \answerTODO{Yes}
     \item Do you justify how the proposed evaluation is sufficient and appropriate to the claims made? 
    \answerTODO{Yes}
     \item Do you discuss what is ``the cost`` of misclassification and fault (in)tolerance?
    \answerTODO{Answer}
  
\end{enumerate}

\item Additionally, if you are using existing assets (e.g., code, data, models) or curating/releasing new assets, \textbf{without compromising anonymity}...
\begin{enumerate}
  \item If your work uses existing assets, did you cite the creators?
    \answerTODO{Yes}
  \item Did you mention the license of the assets?
    \answerTODO{Yes}
  \item Did you work any new assets in the supplemental material or as a URL?
    \answerTODO{Yes}
  \item Did you discuss whether and how consent was obtained from people whose data you're using/curating?
    \answerTODO{NA}
  \item Did you discuss whether the data you are using/curating contains personally identifiable information or offensive content?
    \answerTODO{Yes}
\item If you are curating or releasing new datasets, did you discuss how you intend to make your datasets FAIR (see \citet{fair})?
\answerTODO{Yes}
\item If you are curating or releasing new datasets, did you create a Datasheet for the Dataset (see \citet{gebru2021datasheets})? 
\answerTODO{Yes}
\end{enumerate}

\item Additionally, if you used crowdsourcing or conducted research with human subjects, \textbf{without compromising anonymity}...
\begin{enumerate}
  \item Did you include the full text of instructions given to participants and screenshots?
    \answerTODO{NA}
  \item Did you describe any potential participant risks, with mentions of Institutional Review Board (IRB) approvals?
    \answerTODO{NA}
  \item Did you include the estimated hourly wage paid to participants and the total amount spent on participant compensation?
    \answerTODO{NA}
   \item Did you discuss how data is stored, shared, and deidentified?
   \answerTODO{NA}
\end{enumerate}

\end{enumerate}

\newpage
\appendix

\section{Appendix}
\begin{table*}[!ht]
    \centering
    \small
    \begin{tabular}{|p{1.5cm}lllllll|}
    \hline
        \multicolumn{8}{|l|}{Dataset: SemEval, Event: Hillary Clinton} \\ \hline 
        \textbf{Prompting Scheme} & \textbf{Model name} & \textbf{candidate\_and\_name} & \textbf{misspelling} & \textbf{normal} & \textbf{party} & \textbf{party\_and\_name} & \textbf{underspecify} \\ \hline
        task + context & Falcon-40B instruct-tuned & 0.68 & 0.69 & 0.67 & 0.64 & 0.65 & 0.57 \\ 
        ~ & Falcon-7B instruct-tuned & 0.24 & 0.23 & 0.24 & 0.23 & 0.25 & 0.23 \\ 
        ~ & T5-XL Flan-Alpaca tuned & 0.58 & 0.58 & 0.58 & 0.56 & 0.56 & 0.57 \\ 
        ~ & UL2 Flan-tuned & 0.6 & 0.58 & 0.62 & 0.58 & 0.59 & 0.65 \\ 
        ~ & Llama-2-13b-chat-hf & 0.41 & 0.42 & 0.44 & 0.4 & 0.45 & 0.43 \\ 
        ~ & Llama-2-7b-chat-hf & 0.46 & 0.46 & 0.48 & 0.43 & 0.42 & 0.48 \\ 
        ~ & Mistral-7B instruct-tuned-v0.1 & 0.56 & 0.56 & 0.56 & 0.56 & 0.57 & 0.55 \\ \hline
        zero-shot CoT & Falcon-40B instruct-tuned & 0.53 & 0.52 & 0.49 & 0.47 & 0.5 & 0.51 \\
        ~ & Falcon-7B instruct-tuned & 0.24 & 0.24 & 0.22 & 0.21 & 0.25 & 0.18 \\ 
        ~ & T5-XL Flan-Alpaca tuned & 0.59 & 0.59 & 0.61 & 0.57 & 0.6 & 0.59 \\ 
        ~ & UL2 Flan-tuned & 0.33 & 0.32 & 0.33 & 0.32 & 0.35 & 0.3 \\
        ~ & Llama-2-13b-chat-hf & 0.24 & 0.23 & 0.24 & 0.24 & 0.24 & 0.23 \\ 
        ~ & Llama-2-7b-chat-hf & 0.2 & 0.22 & 0.22 & 0.22 & 0.18 & 0.22 \\ 
        ~ & Mistral-7B instruct-tuned-v0.1 & 0.53 & 0.51 & 0.52 & 0.48 & 0.51 & 0.53 \\ \hline
    \end{tabular}
    \caption{Target Alterations Results for SemEval Dataset}
\end{table*}

\begin{table*}[!ht]
    \centering
    \small
    \begin{tabular}{|p{1.5cm}lllllll|}
    \hline
        \multicolumn{8}{|l|}{Dataset: Election, Event: Hillary Clinton} \\ \hline 
        \textbf{Prompting Scheme} & \textbf{Model name} & \textbf{candidate\_and\_name} & \textbf{misspelling} & \textbf{normal} & \textbf{party} & \textbf{party\_and\_name} & \textbf{underspecify} \\ \hline
        task + context & Falcon-40B instruct-tuned & 0.55 & 0.55 & 0.56 & 0.49 & 0.51 & 0.43 \\ 
        ~ & Falcon-7B instruct-tuned & 0.23 & 0.23 & 0.23 & 0.21 & 0.23 & 0.23 \\ 
        ~ & T5-XL Flan-Alpaca tuned & 0.4 & 0.39 & 0.38 & 0.34 & 0.4 & 0.35 \\ 
        ~ & UL2 Flan-tuned & 0.5 & 0.48 & 0.5 & 0.43 & 0.5 & 0.48 \\ 
        ~ & Llama-2-13b-chat-hf & 0.46 & 0.44 & 0.43 & 0.37 & 0.42 & 0.44 \\ 
        ~ & Llama-2-7b-chat-hf & 0.46 & 0.45 & 0.46 & 0.39 & 0.44 & 0.44 \\ 
        ~ & Mistral-7B instruct-tuned-v0.1 & 0.5 & 0.51 & 0.52 & 0.45 & 0.51 & 0.52 \\ \hline
        zero-shot CoT & Falcon-40B instruct-tuned & 0.45 & 0.45 & 0.44 & 0.4 & 0.44 & 0.41 \\ 
        ~ & Falcon-7B instruct-tuned & 0.27 & 0.28 & 0.26 & 0.27 & 0.28 & 0.25 \\ 
        ~ & T5-XL Flan-Alpaca tuned & 0.5 & 0.48 & 0.49 & 0.45 & 0.49 & 0.48 \\ 
        ~ & UL2 Flan-tuned & 0.32 & 0.31 & 0.31 & 0.23 & 0.32 & 0.31 \\
        ~ & Llama-2-13b-chat-hf & 0.25 & 0.26 & 0.27 & 0.26 & 0.26 & 0.28 \\ 
        ~ & Llama-2-7b-chat-hf & 0.26 & 0.28 & 0.24 & 0.23 & 0.24 & 0.27 \\ 
        ~ & Mistral-7B instruct-tuned-v0.1 & 0.43 & 0.46 & 0.44 & 0.36 & 0.43 & 0.46 \\ 
        ~ & ~ & ~ & ~ & ~ & ~ & ~ & ~ \\ \hline
        \multicolumn{8}{|l|}{Dataset: Election, Event: Donald Trump} \\ \hline 
        \textbf{Prompting Scheme} & \textbf{Model name} & \textbf{candidate\_and\_name} & \textbf{misspelling} & \textbf{normal} & \textbf{party} & \textbf{party\_and\_name} & \textbf{underspecify} \\ \hline
        task + context & Falcon-40B instruct-tuned & 0.51 & 0.47 & 0.53 & 0.42 & 0.41 & 0.37 \\ 
        ~ & Falcon-7B instruct-tuned & 0.28 & 0.28 & 0.27 & 0.27 & 0.3 & 0.27 \\ 
        ~ & T5-XL Flan-Alpaca tuned & 0.31 & 0.3 & 0.28 & 0.27 & 0.3 & 0.27 \\ 
        ~ & UL2 Flan-tuned & 0.43 & 0.41 & 0.43 & 0.39 & 0.45 & 0.39 \\ 
        ~ & Llama-2-13b-chat-hf & 0.43 & 0.43 & 0.41 & 0.39 & 0.43 & 0.44 \\ 
        ~ & Llama-2-7b-chat-hf & 0.38 & 0.35 & 0.35 & 0.37 & 0.39 & 0.35 \\ 
        ~ & Mistral-7B instruct-tuned-v0.1 & 0.44 & 0.47 & 0.46 & 0.39 & 0.43 & 0.48 \\ \hline
        zero-shot CoT & Falcon-40B instruct-tuned & 0.46 & 0.46 & 0.46 & 0.44 & 0.44 & 0.45 \\ 
        ~ & Falcon-7B instruct-tuned & 0.35 & 0.35 & 0.34 & 0.34 & 0.33 & 0.34 \\ 
        ~ & T5-XL Flan-Alpaca tuned & 0.4 & 0.38 & 0.37 & 0.37 & 0.39 & 0.38 \\ 
        ~ & UL2 Flan-tuned & 0.4 & 0.39 & 0.4 & 0.39 & 0.4 & 0.4 \\ 
        ~ & Llama-2-13b-chat-hf & 0.32 & 0.33 & 0.32 & 0.32 & 0.32 & 0.34 \\ 
        ~ & Llama-2-7b-chat-hf & 0.34 & 0.34 & 0.34 & 0.3 & 0.33 & 0.35 \\ 
        ~ & Mistral-7B instruct-tuned-v0.1 & 0.45 & 0.47 & 0.47 & 0.41 & 0.45 & 0.47 \\ \hline
    \end{tabular}
    \caption{Target alterations Results for Elections Dataset}
\end{table*}

\begin{table*}[!ht]
    \centering
    \small
    \begin{tabular}{|p{1.5cm}lllllll|}
    \hline
        \multicolumn{8}{|l|}{Dataset: Basil, Event: Hillary Clinton} \\ \hline 
        \textbf{Prompting Scheme} & \textbf{Model name} & \textbf{candidate\_and\_name} & \textbf{misspelling} & \textbf{normal} & \textbf{party} & \textbf{party\_and\_name} & \textbf{underspecify} \\ \hline
        task + context & Falcon-40B instruct-tuned & 0.68 & 0.39 & 0.36 & 0.63 & 0.74 & 0.28 \\ 
        ~ & Falcon-7B instruct-tuned & 0.39 & 0.43 & 0.43 & 0.38 & 0.39 & 0.38 \\ 
        ~ & T5-XL Flan-Alpaca tuned & 0.27 & 0.26 & 0.27 & 0.24 & 0.32 & 0.29 \\ 
        ~ & UL2 Flan-tuned & 0.19 & 0.18 & 0.18 & 0.2 & 0.18 & 0.15 \\ 
        ~ & Llama-2-13b-chat-hf & 0.16 & 0.22 & 0.14 & 0.2 & 0.19 & 0.12 \\
        ~ & Llama-2-7b-chat-hf & 0.43 & 0.46 & 0.46 & 0.39 & 0.42 & 0.43 \\ 
        ~ & Mistral-7B instruct-tuned-v0.1 & 0.32 & 0.31 & 0.31 & 0.24 & 0.35 & 0.38 \\ \hline
        zero-shot CoT & Falcon-40B instruct-tuned & 0.5 & 0.38 & 0.5 & 0.53 & 0.39 & 0.35 \\ 
        ~ & Falcon-7B instruct-tuned & 0.24 & 0.29 & 0.25 & 0.24 & 0.23 & 0.29 \\ 
        ~ & T5-XL Flan-Alpaca tuned & 0.2 & 0.22 & 0.2 & 0.3 & 0.25 & 0.22 \\ 
        ~ & UL2 Flan-tuned & 0.24 & 0.2 & 0.22 & 0.24 & 0.24 & 0.16 \\ 
        ~ & Llama-2-13b-chat-hf & 0.22 & 0.17 & 0.32 & 0.29 & 0.23 & 0.28 \\ 
        ~ & Llama-2-7b-chat-hf & 0.08 & 0.1 & 0.14 & 0.09 & 0.23 & 0.09 \\ 
        ~ & Mistral-7B instruct-tuned-v0.1 & 0.25 & 0.25 & 0.26 & 0.38 & 0.27 & 0.23 \\ \hline
        \multicolumn{8}{|l|}{Dataset: Basil, Event: Donald Trump} \\ \hline 
        \textbf{Prompting Scheme} & \textbf{Model name} & \textbf{candidate\_and\_name} & \textbf{misspelling} & \textbf{normal} & \textbf{party} & \textbf{party\_and\_name} & \textbf{underspecify} \\ \hline
        task + context & Falcon-40B instruct-tuned & 0.35 & 0.36 & 0.37 & 0.34 & 0.37 & 0.29 \\ 
        ~ & Falcon-7B instruct-tuned & 0.34 & 0.29 & 0.3 & 0.3 & 0.35 & 0.32 \\
        ~ & T5-XL Flan-Alpaca tuned & 0.27 & 0.27 & 0.27 & 0.27 & 0.27 & 0.28 \\ 
        ~ & UL2 Flan-tuned & 0.2 & 0.22 & 0.21 & 0.19 & 0.22 & 0.23 \\ 
        ~ & Llama-2-13b-chat-hf & 0.29 & 0.17 & 0.14 & 0.3 & 0.29 & 0.13 \\ 
        ~ & Llama-2-7b-chat-hf & 0.34 & 0.35 & 0.37 & 0.37 & 0.3 & 0.33 \\ 
        ~ & Mistral-7B instruct-tuned-v0.1 & 0.41 & 0.42 & 0.38 & 0.41 & 0.39 & 0.37 \\ \hline
        zero-shot CoT & Falcon-40B instruct-tuned & 0.29 & 0.39 & 0.38 & 0.29 & 0.32 & 0.34 \\ 
        ~ & Falcon-7B instruct-tuned & 0.23 & 0.24 & 0.21 & 0.33 & 0.25 & 0.27 \\ 
        ~ & T5-XL Flan-Alpaca tuned & 0.28 & 0.29 & 0.31 & 0.27 & 0.3 & 0.24 \\ 
        ~ & UL2 Flan-tuned & 0.24 & 0.22 & 0.21 & 0.19 & 0.24 & 0.15 \\ 
        ~ & Llama-2-13b-chat-hf & 0.23 & 0.21 & 0.17 & 0.24 & 0.23 & 0.2 \\ 
        ~ & Llama-2-7b-chat-hf & 0.13 & 0.12 & 0.14 & 0.11 & 0.12 & 0.14 \\ 
        ~ & Mistral-7B instruct-tuned-v0.1 & 0.37 & 0.36 & 0.31 & 0.27 & 0.37 & 0.3 \\ \hline
    \end{tabular}
    \caption{Target alterations Results for Basil Dataset}
\end{table*}

\end{document}